\documentclass[conference]{IEEEtran}
\IEEEoverridecommandlockouts
\usepackage[dvipdfmx]{graphicx}
\usepackage{amsmath}
\usepackage{latexsym}
\usepackage{ulem}
\usepackage{algorithm}
\usepackage{algpseudocode}
\usepackage{here}
\usepackage{fancyhdr}
\usepackage{fancyvrb}
\usepackage{algorithm}
\usepackage{algpseudocode}
\usepackage{cite}
\usepackage{color}
\usepackage{colortbl}
\usepackage{amsmath}
\usepackage{amsthm}
\usepackage{amssymb}
\usepackage{latexsym}
\usepackage{subfigure}
\usepackage{tikz}
\usepackage{textcomp}
\usepackage{hyperref}
\usepackage{lipsum}
\usepackage{CJKutf8}
\usepackage{rotating}

\newcommand\copyrighttext{%
  \footnotesize \textcopyright 2022 IEEE.  Personal use of this material is permitted.  Permission from IEEE must be obtained for all other uses, in any current or future media, including reprinting/republishing this material for advertising or promotional purposes, creating new collective works, for resale or redistribution to servers or lists, or reuse of any copyrighted component of this work in other works.
  DOI: 10.1109/IIAI-AAI55812.2022.00026}
\newcommand\copyrightnotice{%
\begin{tikzpicture}[remember picture,overlay]
\node[anchor=south,yshift=10pt] at (current page.south) {\fbox{\parbox{\dimexpr\textwidth-\fboxsep-\fboxrule\relax}{\copyrighttext}}};
\end{tikzpicture}%
}

\def\BibTeX{{\rm B\kern-.05em{\sc i\kern-.025em b}\kern-.08em
    T\kern-.1667em\lower.7ex\hbox{E}\kern-.125emX}}
\begin{document}

\title{Developing a Component Comment Extractor\\from Product Reviews on E-Commerce Sites}

\author{\IEEEauthorblockN{Shogo Anda, Masato Kikuchi, Tadachika Ozono}
\IEEEauthorblockA{\textit{Department of Computer Science, Graduate School of Engineering} \\
\textit{Nagoya Institute of Technology}\\
Gokiso-cho, Showa-ku, Nagoya, Aichi, 466-8555, Japan \\
Email: anda@ozlab.org, \{kikuchi, ozono\}@nitech.ac.jp}}

\maketitle

\copyrightnotice
\begin{abstract}
    Consumers often read product reviews to inform their buying decision,
    as some consumers want to know a specific component of a product.
    However, because typical sentences on product reviews contain various details,
    users must identify sentences about components they want to know amongst the many reviews. 
    Therefore, we aimed to develop a system that identifies and collects component and aspect information of products 
    in sentences.
    Our BERT-based classifiers assign labels referring to components and aspects to sentences in reviews
    and extract sentences with comments on specific components and aspects.
    We determined proper labels based for the words identified through pattern matching from product reviews 
    to create the training data.
    Because we could not use the words as labels, 
    we carefully created labels covering the meanings of the words.
    However, the training data was imbalanced on component and aspect pairs.
    We introduced a data augmentation method using WordNet to reduce the bias.
    Our evaluation demonstrates that 
    the system can determine labels for road bikes using pattern matching, 
    covering more than 88\% of the indicators of components and aspects on e-commerce sites.
    Moreover, our data augmentation method can improve 
    the \boldmath{$macro\mathchar`-F1\mathchar`-measure$} on insufficient data from 0.66 to 0.76.
\end{abstract}

\begin{IEEEkeywords}
information extraction, BERT, data augmentation, product review.
\end{IEEEkeywords}

\section{Introduction}
When purchasing products online,
we gather information from reviews to help inform our choice,
particularly in products composed of multiple components, such as bicycles and PCs.
Reviews include mixed comments on each component of the product.
Some users require a support system to find comments on specific components,
as manually collecting comments on specific components requires considerable time and effort.

We aimed to develop a comment extractor that extracts individual components from reviews.
For a sentence in a road bike review such as ``Tires went flat immediately'',
we assign ``Tire'' as the component label and ``Durability'' as the aspect label 
and extract the sentence as a comment on the specific component, the tire, and specific aspect, the durability.
Two tasks achieve this purpose.
The first is the difficult creation of training data for use during supervised classifier training.
Creating training data requires some defined classification labels.
However, using the different product components and aspects of each component as classification labels is difficult.
Some products have component or aspect indicators on e-commerce sites that are usable as labels, but not all do, 
and products that are not bought or sold component by component often lack these indicators.
Therefore, designing labels that consider the different components and aspects of each product type is necessary.
The second is the creation of label pairs indicating components and aspects will cause bias within the training data.
For example, comments on the durability of tires appear more often than the functionality of tires in road bike reviews.
Training with biased data is known to degrade the classification performance of trained classifiers\cite{Lopez}.
Therefore, reducing the degradation of classification performance is necessary when training with biased data.

We present a method for assigning component and aspect labels mentioned in a sentence using two BERT-based\cite{BERT} classifiers.
We also describe our two approaches for the two tasks.
The first approach uses pattern matching to create classification labels from reviews for training data
 that considers the differences in components and aspects by product type.
We focus on the occurrence pattern of component names and aspect words in Japanese sentences,
 automatic extraction of component names and aspect words to create each label through pattern matching,
 and manually creating each label from these.
The second approach involves data augmentation by generating similar sentences to reduce bias in the training data.
We generated similar sentences through synonym replacement using WordNet\cite{WordNet}
 and added to the training data to augment the small amount of data.

We evaluated the performance of our label-creating method and effect of our data augmentation method.
Our label-creating method matches 88\% of the e-commerce site indicators of components and aspects,
and our data augmentation method improves classification performance up a $macro\mathchar`-F1\mathchar`-measure$ of 0.78.
From our results, we conclude that our label-creating and data augmentation methods effectively assign components and aspects to sentences based on BERT.

The remainder of this paper is organized as follows. First,
 we show related work in Section II.
 Then, Section III explains our comment extraction method and its training data,
 and Section IV described our evaluation of our method for component comment extraction.
 Finally, we discuss the system in Section V and conclude our results in Section VI.
\section{Related Work}
Various methods for analyzing reviews have been examined by many researchers
in the field of natural language processing.
Haque et al.\cite{Haque} attempted to classify the sentiment of reviews
using a supervised learning model.
In addition, Xu et al.\cite{Xu} attempted to use product reviews as resources for answering questions about the product.
In contrast, this study attempts to classify sentences in product reviews
 that focus on the individual components that compose the product and aspects mentioned.  

In this study, our application domain is Japanese e-commerce sites.
Japanese text does not use spaces between words,
requiring creative extraction methods for component names and aspect words.
Kobayashi et al.\cite{kobayashi2} determined that
opinions in Japanese text are composed of the writer and object (object),
the component and attribute to which the object belongs (aspect),
and a positive or negative sentiment (sentiment).
In particular, they focused on the extraction of aspect-sentiment and aspect-of relations.
In their experiments, they showed that a context-aware model is effective
in extracting these relations in Japanese.

We extract words corresponding to components in products 
and words related to evaluations from the reviews
and label sentences based on those words to create training data that
accounts for the different components and aspects of each product type.
In label creation, in addition to the patterns used by Kobayashi et al., 
frequent patterns in product reviews were extracted for use as the basis for creating labels.

We augmented the training data with sentences.
For the augmentation, we used synonym replacement,
 which was introduced as a simple and easy data augmentation by Wei et al.\cite{EDA}. 
Synonyms were obtained using WordNet.

\section{Extracting Comments on Each Component}
This section explains our comment extraction system 
that finds product component and aspect information in a sentence. 
The system consists of two BERT-based classifiers; a component classifier and an aspect classifier. 
The component (or aspect) classifier determines the component (or aspect) labels for each review sentence. 
For example, our system may find a sentence such as ``The tire went flat immediately.'' The sentence contains ``tire'' as a product component and ``went flat immediately'' as its aspect. 
First, we describe how to determine component and aspect label sets in \ref{A}. 
Second, \ref{B} shows our classifier models based on BERT. 
Finally, in \ref{C}, we explain our dataset augmentation. 
\begin{table}[h]
    \caption{Component label candidates. Candidates may contain incorrect labels such as ``assembly''.}
    \label{RK}
    \begin{center}
        \begin{tabular}{l|r}
            \hline
            Component name&Number of occurrences\\ \hline
            assembly (improper)&250\\
            bicycle (improper)&184\\
            brake&153\\
            tire&145\\
            bike&84\\
            saddle&75\\
            pedal&75\\
            price (improper)&69\\
            bicycle shop (improper)&55\\
            handle&54\\
            \hline
        \end{tabular}%
    \end{center}
\end{table}
\begin{table}[t]
    \caption{Extracted aspect words}
    \vspace{-3mm}
    \label{AK}
    \begin{center}
        \begin{tabular}{l|l}
            \hline
            Component label&Aspect words\\
            \hline
            Brake&\textsl{\begin{CJK}{UTF8}{ipxm}効く\end{CJK}} (effect), \textsl{\begin{CJK}{UTF8}{ipxm}弱い\end{CJK}} (weak), \textsl{\begin{CJK}{UTF8}{ipxm}甘い\end{CJK}} (weak)\\
            Tire&\textsl{\begin{CJK}{UTF8}{ipxm}パンク\end{CJK}} (got flat), \textsl{\begin{CJK}{UTF8}{ipxm}細い\end{CJK}} (narrow), \textsl{\begin{CJK}{UTF8}{ipxm}歪む\end{CJK}} (distorted)\\
            Saddle&\textsl{\begin{CJK}{UTF8}{ipxm}硬い\end{CJK}} (hard), \textsl{\begin{CJK}{UTF8}{ipxm}堅い\end{CJK}} (hard)\\
            Pedal&\textsl{\begin{CJK}{UTF8}{ipxm}外れる\end{CJK}} (loose), \textsl{\begin{CJK}{UTF8}{ipxm}ぐらつく\end{CJK}} (unsteady), \textsl{\begin{CJK}{UTF8}{ipxm}重い\end{CJK}} (heavy)\\
            Handle&\textsl{\begin{CJK}{UTF8}{ipxm}曲がる\end{CJK}} (misalignment)\\
            \hline
        \end{tabular}%
    \end{center}
    \vspace{-2mm}
\end{table}
\begin{table}[t]% Table 2
    \caption{Component labels for road bikes}
    \vspace{-3mm}
    \label{RG}
    \begin{center}
    \begin{tabular}{c|c}
    \hline
    &Labels\\\hline
    Component labels&Tire, Valve, Rim, Spoke, Handle, Brake, Bell, Gear,\\
    (14 labels)&Pedal, Crank, Chain, Light, Saddle, Frame\\
    \hline
    Aspect labels&Durability, Functionality, Preference, Installation,\\
    (7 labels)&Weight, Size, Appearance\\
    \hline
    \end{tabular}
    \vspace{-3mm}
    \end{center}
  \end{table}

\subsection{Label-Creating Method}\label{A}
This section describes how to determine component and aspect label sets 
from product reviews to create training datasets. 
E-commerce sites provide some labels, for example, 
``Tire'' as a component label and ``Durability'' as an aspect label, 
but they are insufficient for our purpose. 
Therefore, we must develop a method for identifying more proper labels from product reviews. 
Creating the label sets consists of three parts;
 1) making sentence patterns to find sentences possibly containing label candidates,
 2) collecting label candidates by applying patterns to reviews,
 and 3) creating proper labels manually. 
Subsequently, we created training datasets for the two classifiers 
for component and aspect classification.

First, we describe the patterns used to collect the label candidates. 
Our patterns were;
 1) $\langle component \rangle$ ``\begin{CJK}{UTF8}{ipxm}の\end{CJK}'' $\langle aspect \rangle$ ($\langle aspect\rangle$ of $\langle component \rangle$)
 and 2) $\langle component \rangle$ ``\begin{CJK}{UTF8}{ipxm}が\end{CJK}'' $\langle aspect \rangle$ ($\langle component \rangle$ is $\langle aspect \rangle$).
In Japanese sentences, evaluative expressions appear in the form\cite{Ekobayashi}:
\vspace{-2mm}
\begin{center}
    \vspace{-6mm}
    \begin{align*}\label{form}
        \langle component \rangle \textsl{\begin{CJK}{UTF8}{ipxm}の\end{CJK}} \langle aspect \rangle \textsl{\begin{CJK}{UTF8}{ipxm}は\end{CJK}} \langle sentiment \rangle \cdots \text{Japanese}
        \vspace{-10mm}
    \end{align*}
    \vspace{-1mm}
    ($\langle aspect \rangle$ of $\langle component \rangle$ is $\langle sentiment \rangle$ $\cdots$English).
\end{center}
In addition, as shown below, 
component names appear as $\langle component \rangle$ in a review sentence,
and aspect words appear as $\langle aspect \rangle$ \hspace{-1mm}.
In addition to the form,
 component names and aspect words also appear in review sentences 
 in the form where $\langle component \rangle$ and $\langle aspect \rangle$, which are combined by ``\textsl{\begin{CJK}{UTF8}{ipxm}が\end{CJK}}''.
They appear as \textsl{``\begin{CJK}{UTF8}{ipxm}ブレーキ\end{CJK}␣\hspace{0.4mm}\begin{CJK}{UTF8}{ipxm}の\end{CJK}␣\hspace{0.4mm}\begin{CJK}{UTF8}{ipxm}効き\end{CJK}␣\hspace{0.4mm}\begin{CJK}{UTF8}{ipxm}が\end{CJK}␣\hspace{0.4mm}\begin{CJK}{UTF8}{ipxm}悪い\end{CJK}''} (Brakes are awful), 
\textsl{``\begin{CJK}{UTF8}{ipxm}タイヤ\end{CJK}␣\hspace{0.4mm}\begin{CJK}{UTF8}{ipxm}が\end{CJK}␣\hspace{0.4mm}\begin{CJK}{UTF8}{ipxm}パンク\end{CJK}␣\hspace{0.4mm}\begin{CJK}{UTF8}{ipxm}した\end{CJK}''} (Tires went flat), 
and \textsl{``\begin{CJK}{UTF8}{ipxm}スポーク\end{CJK}␣\hspace{0.4mm}\begin{CJK}{UTF8}{ipxm}が\end{CJK}␣\hspace{0.4mm}\begin{CJK}{UTF8}{ipxm}折れた\end{CJK}''} (Spokes got broken).

Second, we collected label candidates. First, we collected 30,000 sentences 
from product reviews in the categories
of Sports and Outdoors, Cycling, Bikes, and Road Bikes
on Amazon.com. Next, we separated sentences into words 
using a Japanese morphological analyzer MeCab\footnote{https://taku910.github.io/mecab/}. Finally, we collected label candidates from the sentences matched to the patterns.

TABLE \ref{RK} lists the 10 most frequent component names in the candidates. 
We had to remove words such as ``assembly'' and ``price'' 
because these are not component names. 
In addition, words such as ``bicycle'' and ``bicycle shop'' are inappropriate for component labels.

As shown in TABLE \ref{AK}, 
we found that aspect words include many varieties of expression and notation, 
such as ``\textsl{\begin{CJK}{UTF8}{ipxm}弱い\end{CJK}}'' (weak) and ``\textsl{\begin{CJK}{UTF8}{ipxm}甘い\end{CJK}}'' (weak) for brakes, 
``\textsl{\begin{CJK}{UTF8}{ipxm}硬い\end{CJK}}'' (hard) and ``\textsl{\begin{CJK}{UTF8}{ipxm}堅い\end{CJK}}'' (hard) for the saddle. 
Therefore, we had to determine aspect labels that consider these expressions.

Finally, we manually created component and aspect labels on road bikes, as shown in TABLE \ref{RG}.
We determined the aspect labels that consider the meanings of extracted aspect words instead of using the extracted words.
We created 14 and 7 labels for components and aspects respectively.

The sentences in our training data have labels in the two label sets. 
For example, the sentence ``the spoke is quite heavy'' can have a component label ``Spoke'' 
and aspect label ``Weight''. 
We created a training dataset with 1,000 sentences using these labels.

\subsection{Classifier Model}\label{B}
This subsection describes the classifier model
that assigns two types of labels to sentences.

We created two classifiers, the component classifier and aspect classifier, based on BERT.
As the target of this study is Japanese sentences,
we used the Japanese pretrained model (cl-tohoku/bert-base-japanese)
available from HuggingFace\footnote{https://huggingface.co}.

Each classifier takes a sentence as input
and outputs component labels for the input sentence in the component classifier
and aspect labels in the aspect classifier.
They have four layers.
The BERT Tokenizer splits the sentence into word units and prefixes those words with the [CLS] token, to use as input for BERT.
When using BERT for classification tasks, we use the vector regarding the [CLS] token.
The output of BERT regarding the [CLS] token is input to Linear,
where the dimension of the output is changed to the number of each label,
and then converted by Softmax such that the number of each dimension has a closed interval value of [0,1].
For the output of the two classifiers,
each value is converted to zero or one using a predefined threshold value,
and a label corresponding to the dimension with a value of one is assigned to the input sentence.

In this manner, we used trained classifiers to assign two types of labels to the input sentences.
\subsection{Data Augmentation Method}\label{C}
This subsection describes the data augmentation
with similar sentence generation
to reduce bias in the training data.

Two types of labels were manually assigned to sentences in the review to create training data.
Training data was biased because of the component and aspect pairs.
For example, some components were biased by being included more as a pair with a particular aspect
and less as a pair with another aspect.
TABLE \ref{RD} breaks down the data
used in the aforementioned specific example of label creation,
in which two labels were manually assigned to 1,000 of the 30,000 sentences collected.
Each number in TABLE \ref{RD} is the number of the sentences in the dataset in which the corresponding pair is mentioned.
\begin{table}[t]% Table 2
	\centering
  \caption{labeled data breakdown}
  \label{RD}
  \vspace{2mm}
  \begin{tabular}{l|rrrrrrr|r}
  \hline
%   &\pbox<t>{Durability}&\pbox<t>{Functionality}&\pbox<t>{Preference}&\pbox<t>{Installation}&\pbox<t>{Weight}&\pbox<t>{Size}&\pbox<t>{Appearance}&Total\\
  &\begin{turn}{-90}Durability\end{turn}&\begin{turn}{-90}Functionality \hspace{2mm}\end{turn}&\begin{turn}{-90}Preference\end{turn}&\begin{turn}{-90}Installation\end{turn}&\begin{turn}{-90}Weight\end{turn}&\begin{turn}{-90}Size\end{turn}&\begin{turn}{-90}Appearance\end{turn}&Total\\
  \hline
  Tire&70&5&22&53&3&23&22&170\\
  Valve&8&0&19&2&0&2&2&33\\
  Rim&15&1&10&19&0&5&15&51\\
  Spoke&19&0&0&5&0&0&2&21\\
  Handle&33&11&20&73&8&11&14&137\\
  Brake&27&21&30&85&4&1&8&147\\
  Bell&9&1&11&2&0&0&6&25\\
  Gear&24&37&86&41&6&2&3&163\\
  Pedal&38&5&23&68&7&2&10&117\\
  Crank&25&4&7&20&1&1&0&42\\
  Chain&38&12&7&35&1&3&13&90\\
  Light&23&1&17&15&0&1&9&82\\
  Saddle&17&6&85&45&3&13&13&151\\
  Frame&22&6&24&23&18&14&55&130\\
  \hline
  Total&283&82&297&322&45&69&145&\\
  \hline
  \end{tabular}
  \vspace{3mm}
\end{table}

\begin{figure}[t]
    \centering
    \vspace{3mm}
    \includegraphics[keepaspectratio, width=\hsize]{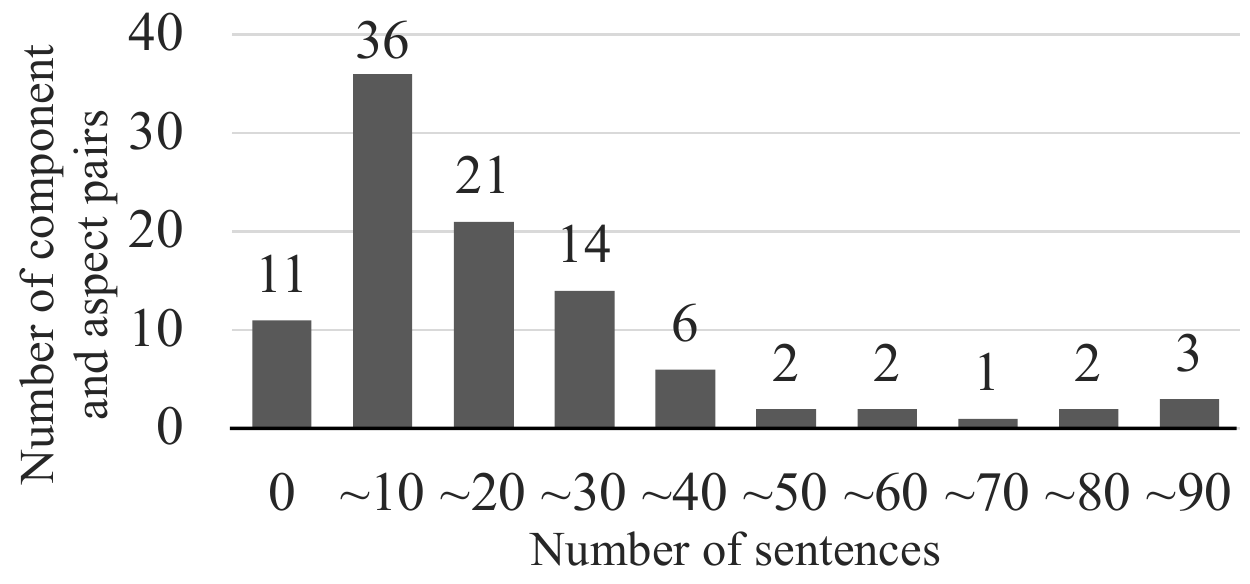} 
    \caption{Histogram showing bias in the number of sentences in each pair.}
    \label{HG}
\end{figure}

Fig. \ref{HG} shows the histogram of TABLE \ref{RD}.
As seen in the histogram,
 the entire histogram is biased to the left,
 as 47 of the total 98 pairs have 10 or fewer sentences,
 and some pairs have more than 80.

Classifier training with such biased data degrades the performance of the trained model.
We reduce bias in the training data by augmenting the data with synonym replacing.
Synonyms were obtained using WordNet.
\begin{figure}[t]
    \centering
    \includegraphics[keepaspectratio, width=\hsize]{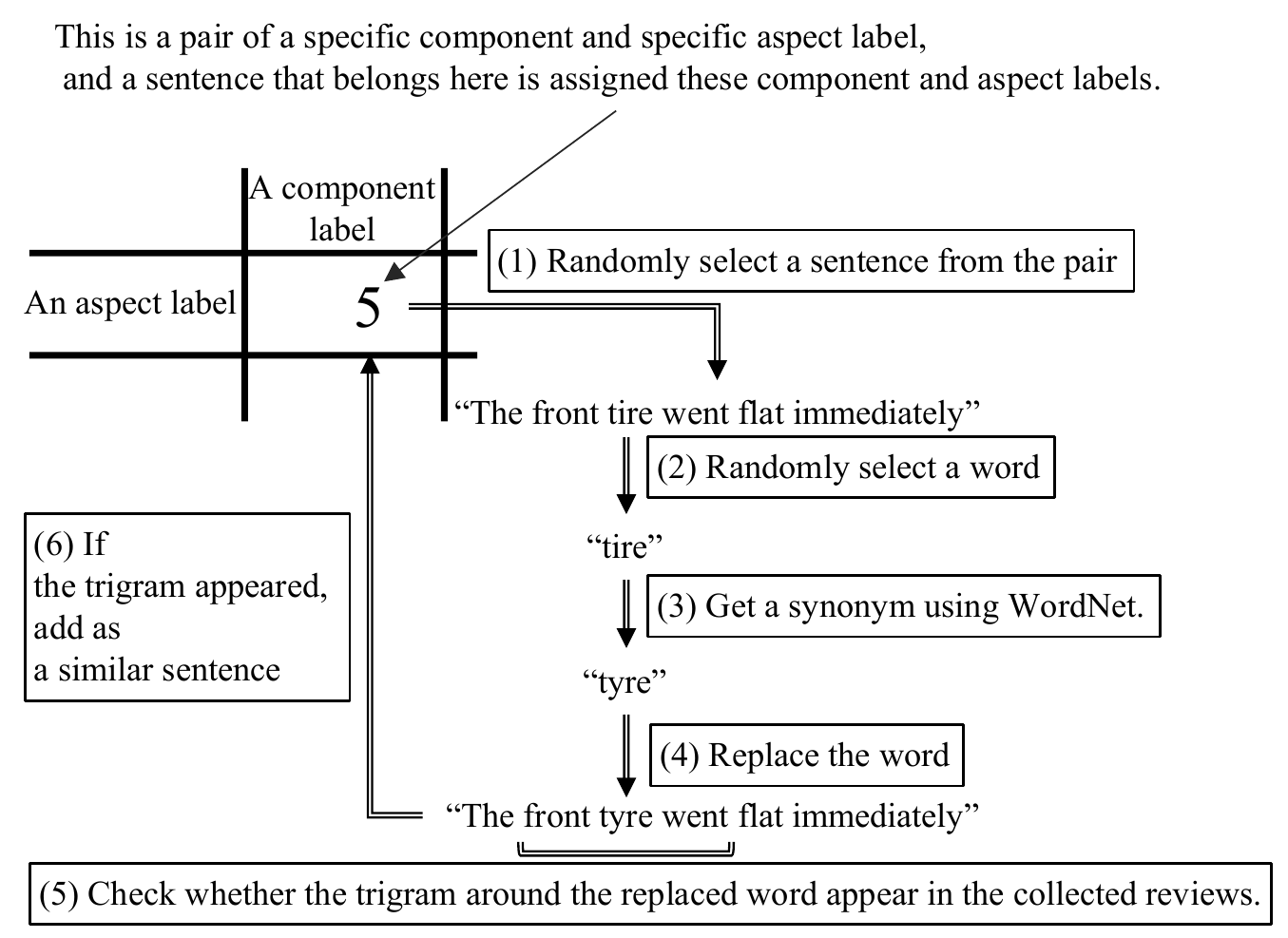} 
    \caption{Process for adding similar sentences to pairs that do not satisfy the Minimum Pair Size.}
    \label{DA}
\end{figure}
In this study, pairs in the training data with three or more sentences were considered.
When augmenting the data, the minimum pair size (MPS) was defined,
and the data was augmented until all pairs
considered have sentences greater than or equal to the MPS.
Fig. \ref{DA} presents the process of similar sentence generation during data augmentation.
This flow exemplifies the generation of similar sentences 
when a component and aspect label pair have five sentences and are less than the MPS.
First, we randomly selected a sentence from the pair (1) and divided the sentence into word units.
 We used the Python module termextract to perform the division, 
 considering compound words and technical terms. 
Next, we randomly selected one word from the words (2). 
 We obtained synonyms of the selected word using WordNet (3). 
 We replaced the synonym with the original word to generate similar sentences (4). 
Finally, we took the three words in the generated sentences,
 the replaced word, and the words before and after it
 as the trigram and check if they appear in the collected reviews (5).
 If that trigram appeared,
 we add the generated similar sentence to the training data (6). 
 If it does not appear, we return to random sentence selection (1).
The similar sentence is given the same label as the source sentence and added to the training data.
We augmented the training data by repeating the process
 until the number of sentences equals the MPS for all pairs.
If the sentences generated by checking the trigram do not satisfy the MPS,
the check is changed to a bigram consisting of two words,
 the replaced word and word before or after it,
 and then the unigram with only the replaced word to augment the data.

 Thus, our data augmentation using similar sentence generation reduces bias in the training data.
\begin{figure}[t]
    \centering
    \includegraphics[keepaspectratio, width=\hsize]{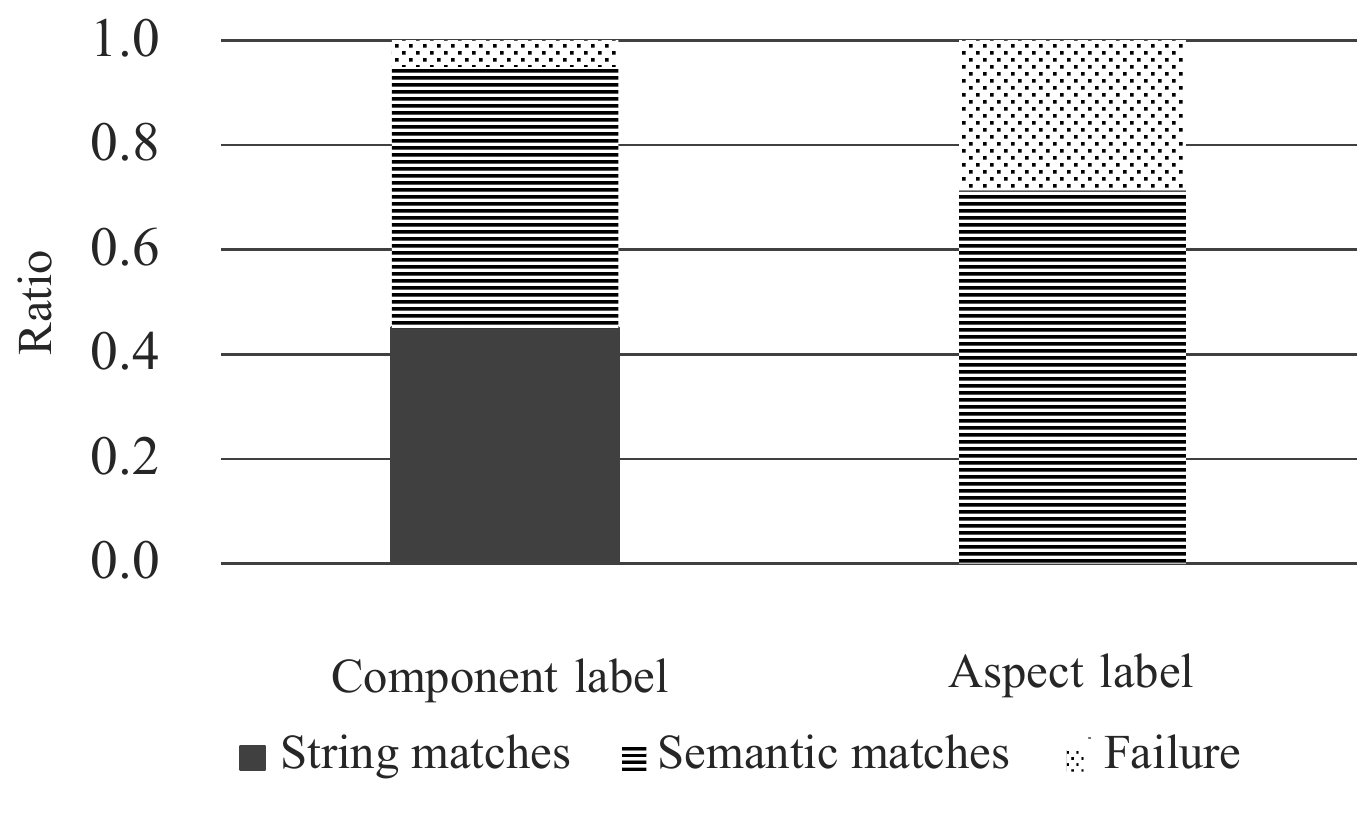} 
    \caption{Results of evaluation Experiment 1}
    \label{H1}
\end{figure}
\begin{table}[tb]
    \centering
    \caption{Dataset Details}
    \label{DS}
    \begin{tabular}{l|r|r|r|r|l}
    \hline
    Type&Size&
    \begin{tabular*}{13mm}{@{\extracolsep{\fill}}c}
        Component\\labels\\
    \end{tabular*}&
    \begin{tabular*}{8mm}{@{\extracolsep{\fill}}c}
    Aspect\\labels\\
    \end{tabular*}&Pairs&
    \begin{tabular}{c}
        Expected\\classification\\difficulty\\
    \end{tabular}\\
    \hline
    Road bike&500&14&7&50&baseline\\
    Laptop PC&500&10&7&47&more difficult\\
    Tent&500&10&6&38&easier\\
    \hline
    \end{tabular}
    \vspace{-1mm}
\end{table}
\begin{figure}[t]
    \centering
    \includegraphics[keepaspectratio, width=\hsize]{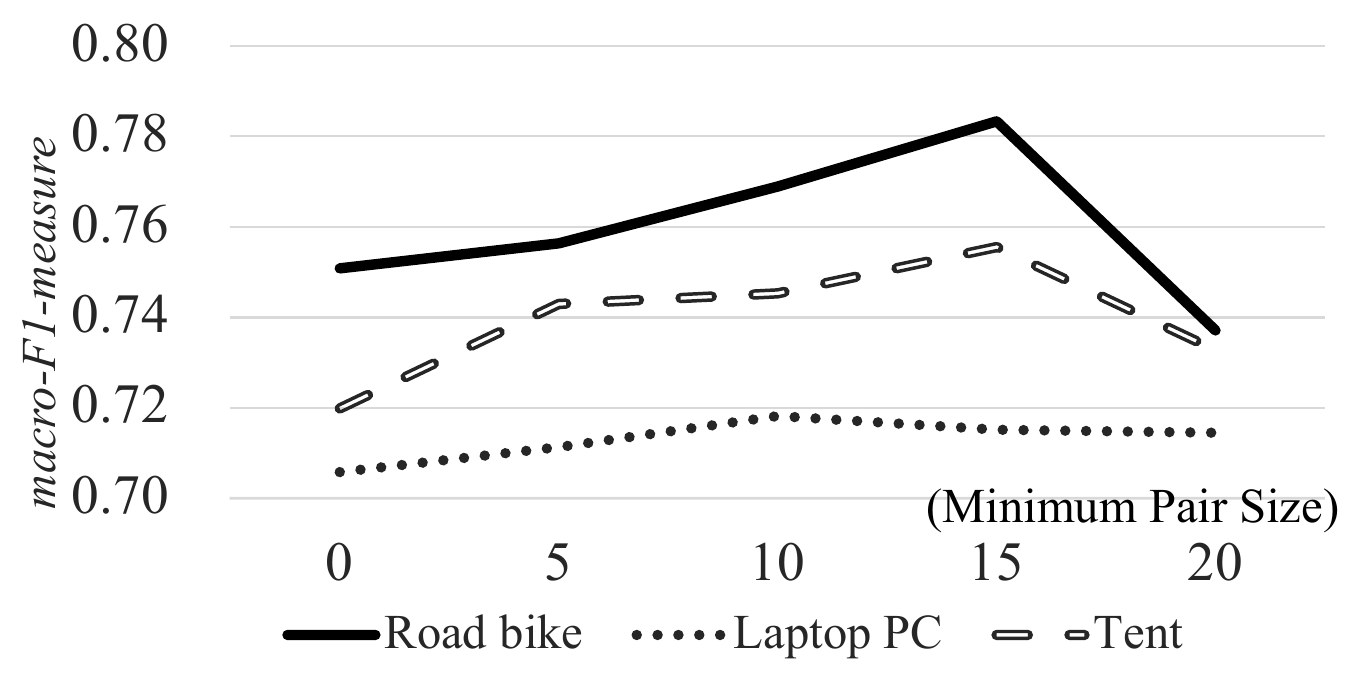} 
    \vspace{-5mm}
    \caption{Results of evaluation Experiment 2}
    \label{H2}
\end{figure}
\section{Experiments and Results}
This section describes the settings and results of two experiments
that evaluated our label-creating and data augmentation methods.
In evaluation Experiment 1,
 we assessed whether our label-creating method can create labels of the components and aspects 
 mentioned in the reviews.
We used a dataset of road bike reviews with component and aspect indicators 
on an e-commerce site to evaluate the percentage match percentage 
between the labels and indicators on the e-commerce site. 
We also evaluated the effect of our data augmentation method 
on classification performance by calculating the relationship 
between the MPS set during data augmentation and that during classification performance.

\subsection{Evaluation Experiment 1}
This subsection describes the setting and result of Experiment 1, in which the label-creating method was evaluated.

We compared the labels using our method to the component and aspect indicators on the e-commerce site.
Component labels were compared with categories of bike components and components on Amazon,
and aspect labels were compared with evaluation items for road bikes on Kakaku.com\footnote{https://review.kakaku.com/}.
Kakaku.com is a site that compares products of the same genre on various e-commerce sites,
 and, depending on the product type, 
 several evaluation items for that product are manually defined.
To evaluate our label-creating method, we used road bikes that have a components category on Amazon and defined evaluation items on Kakaku.com.
\begin{figure*}[t]
    \centering
    \includegraphics[keepaspectratio, width=\hsize]{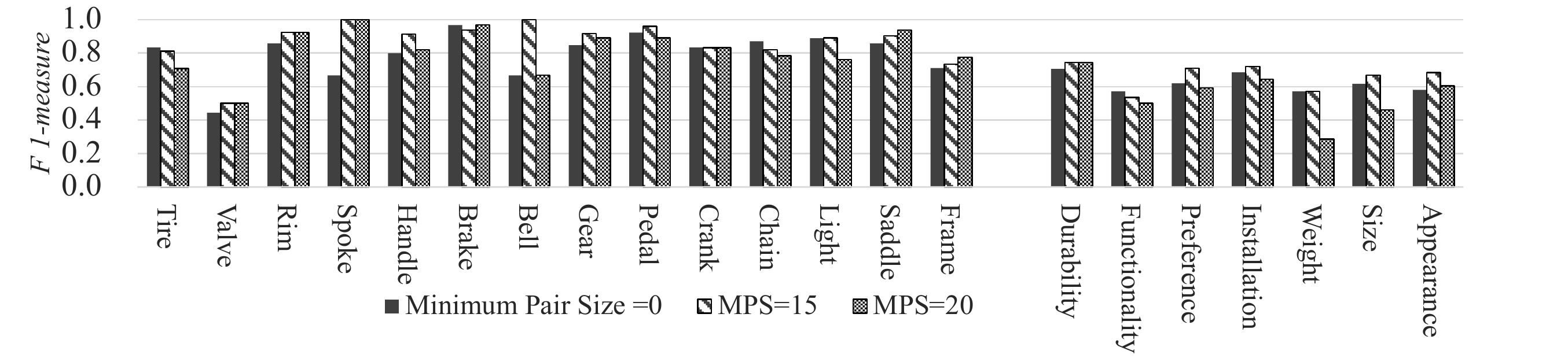} 
    \caption{$F1\mathchar`-measure$s for each label in Road Bikes. The left side is the component label and the right side is the aspect label.}
    \label{H2-2}
\end{figure*}
\subsubsection{Experimental Settings}
After collecting 10,000 sentences on road bike reviews from Amazon
and extracting component names and aspect words through pattern matching,
we manually created two types of labels based on the component names and aspect words.
The 14 component labels and 7 aspect labels were compared to the indicators on the e-commerce site.

For comparison,
in the component category on Amazon,
categories that match the component label with a string are considered string matches,
and, in the two indicators,
categories that match the label by considering expression and notation variabilities 
are considered semantic matches.
Components not mentioned in the review,
such as tires and tire tubes and brakes and brake wires,
are integrated based on the product categories in Cycle Base Asahi\footnote{https://ec.cb-asahi.co.jp/category/},
and categories that match the component labels through integration are also considered semantic matches.
Cycle Base Asahi is an online shopping site that sells bicycle products and components.
Its product categories include one for bicycle components,
grouping similar components such as saddles and seat posts.
The semantic match is determined by referring to these categories.
\subsubsection{Experimental Results}
Fig. \ref{H1} shows the results of evaluation Experiment 1.

For component labels and categories,
45\% of component categories were string matches
and 50\% semantic matches.
The total percentage of matches was 95\%.

Aspect labels and evaluation items in Kakaku.com
resulted in semantic matches for 71\% of the evaluation items.
The disagreeing items were ``Driving performance'' and ``Parts.''
``Driving performance" concerns speed and braking performance,
and ``Parts'' concerns the performance of equipped components.
\subsection{Evaluation Experiment 2}
In this experiment,
to evaluate data augmentation by generating similar sentences,
we calculated the effect of the MPS set up 
for training data augmentation
on classification performance.
\subsubsection{Datasets}
We used three datasets in this experiment.
In addition to the road bikes treated as examples in this paper,
we used laptop PCs,
which we estimated to more components than road bikes and thereby more difficult to classify, 
and tents, which we estimated to have less components than road bikes and
thereby easier to classify.

TABLE \ref{DS} lists the details of the datasets.
We manually assigned two labels to the sentences in each sentence in the three datasets.

Among the three datasets, 
we randomly selected two sentences from each pair with five or more sentences
and used them as evaluation data.
In addition, we randomly selected 50 sentences excluded from the evaluation data to use as validation data,
and the remainder as training data.
\subsubsection{Evaluation Scale}
We computed the performance metrics using $Precision_x$, $Recall_x$.
These are values derived from whether the assigned and true labels have label $x$.
$F1\mathchar`-measure_x$ is given by
\begin{align*}
    F1\mathchar`-measure_x = 2\cdot\frac{Precision_x\cdot Recall_x}{Precision_x+Recall_x}.
\end{align*}
The $macro\mathchar`-F1\mathchar`-measure$ is the macro average of labels,
\begin{align*}
    macro\text{-}F1\text{-}measure =\frac{1}{|Labels|}\sum_{l\in Labels}F1\text{-}measure_l,
\end{align*}
where $Labels$ is the set of labels in the product type, and $|Labels|$ is the size of the $Labels$ set.
\subsubsection{Experimental Settings}
MPS during data augmentation was $\rm{MPS}=0,5,... ,20$.
For the training data in each dataset,
data augmentation was performed such that each item size was greater than or equal to MPS.
For each MPS, the training data was augmented and the model trained.
The validation data was used to set the threshold for converting each value of the output of the model,
which is a vector with real values in [0,1], to zero or one.
The $macro\mathchar`-F1\mathchar`-measure$ of the model in each MPS was calculated using the evaluation data.
\subsubsection{Experiment Results}
Fig. \ref{H2} shows the results of evaluation Experiment 2.

In all three datasets, $F1\mathchar`-measure$s increased in the range of MPS=5, 10, and 15
compared to before data augmentation.
For road bikes, the results are lower than before data augmentation when MPS=20.
Similarly, for tents, the $F1\mathchar`-measure$ is lower for MPS=20 than for the model with MPS=15.
In laptop PCs, the increase in the $F1\mathchar`-measure$ is less than in the other two products.
However, the decrease at MPS=20 observed in the other two datasets is not observed in laptop PCs.

Fig. \ref{H2-2} shows the $F1\mathchar`-measure$ for each label with MPS = 0, 15, and 20 for road bikes.
In labels with fewer data, such as spokes and bells,
the data augmentation temporarily increases the $F1\mathchar`-measure$s,
except in the case of MPS=20, where the $F1\mathchar`-measure$s decrease in some of those labels.

\begin{figure}[t]
    \centering
    \includegraphics[keepaspectratio, width=\hsize]{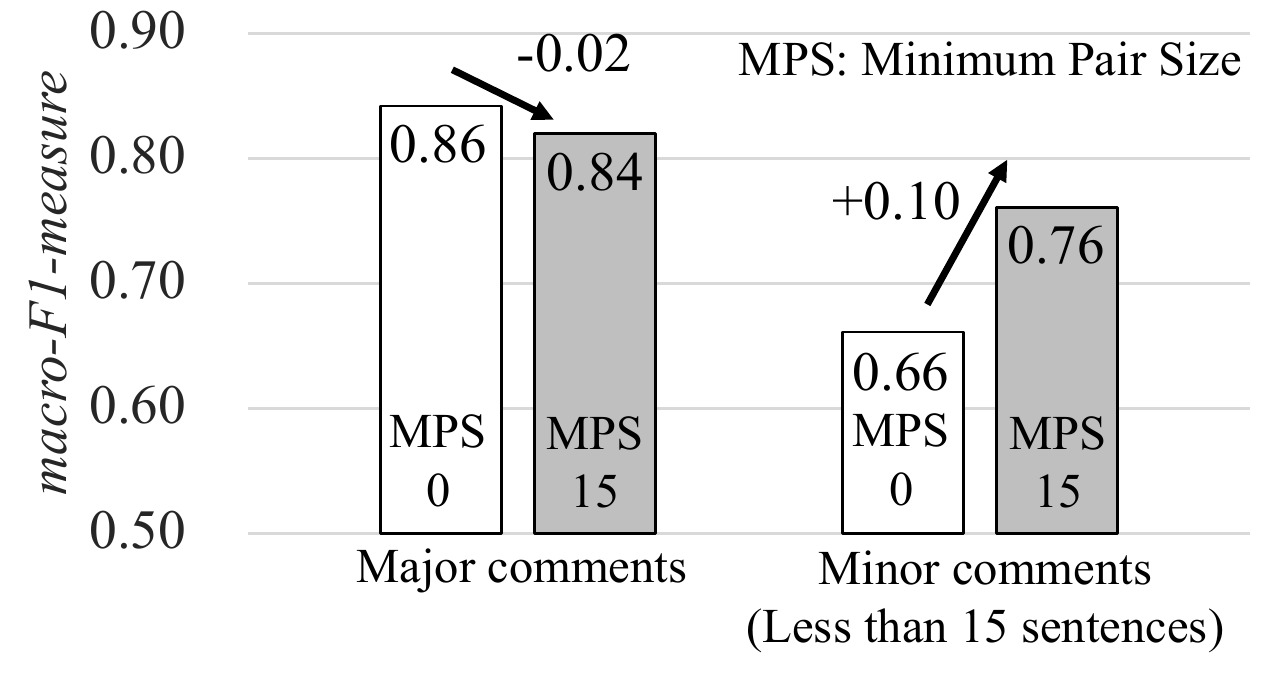} 
    \vspace{-5mm}
    \caption{Comparison between major and minor comments}
    \label{changes}
    \vspace{-2mm}
\end{figure}

We evaluated the differences in classification performance for augmented pairs to assess the effect of our data augmentation method on the small amount of data.
Fig. \ref{changes} shows the differences in $macro\mathchar`-F1\mathchar`-measure$s for pairs with and without data augmentation for major and minor comments.

The $macro\mathchar`-F1\mathchar`-measure$ increased by 0.1 with data augmentation.

\section{Discussion}
The results of evaluation Experiment 1 show that
the pattern matching-based label-creating can make labels consistent 
with most indicators except for the indicator related to ``performance.''
The reason for this is thought to be that
the aspect words related to ``performance'' are subdivided
into ``Durability'' and ``Functionality,'' which become their labels.
In road bike reviews, this method is thought to further subdivide indicators lumped together by ``performance'' on e-commerce sites and assign detailed aspects to sentences.
This type of inclusive relationship between labels is observed in many product types.
For example, wheels, tires, and rims in bicycles also relate to each other.
Although creating labels that consider such relationships is necessary 
among components and aspects,
the current method of manually making labels based on component names and aspect words 
still requires time and effort.
The consideration of relationships between labels and further semi-automation in the label-creating can be investigated in future works.

Our results show that our data augmentation method using similar sentence generation 
can improve the classification performance on almost minor comments in the road bike domain. 
Furthermore,
we found that this augmentation improves the recall rate of the classification. 
However, we could not improve the performance in the laptop PC domain 
because the domain has more technical terms and product and component names than the road bike domain. 
In addition, WordNet does not cover all synonyms in all domains. 
Therefore, we must develop a system extracting domain knowledge, 
including named entities such as product names, on synonyms in a target domain.
This result shows that MPS = 15 is appropriate for a dataset size of 500 in the domain of road bikes and tents.
The optimal MPS can be determined experimentally.

Our data augmentation method improves the performance for small amounts of data in the training dataset. 
The system can also extract minor comments from reviews. 
Extracting minor comments is more beneficial than extracting major comments because humans can already extract major comments.
Moreover, providing extracted comments with their ratings can be useful. 

In addition, this study focused on components and aspects,
and sentiment identification is planned for future work. 
This study aimed to collect information for each component by extracting comments for each component, 
but in the future, considering sentiment for an aspect can lead to, 
for example, predicting the ratings of the components of a product.
\section{Conclusion}
We developed a comment extraction system designed for components and aspects. 
As the system determines labels indicating components and aspects to sentences in the review,
it can extract sentences as comments on specific components and aspects. 
First, we created training data using simple pattern matching from product reviews. 
However, the training data was imbalanced, and some component-aspect pairs lacked sufficient data. 
Therefore, we introduced a data augmentation method that generates similar sentences based on WordNet-based synonym replacement to reduce the bias. 
Finally, our label-creating method can find 88\% of labels on road bikes of the component and aspect indicators on e-commerce sites.
Moreover, our data augment method can improve the $macro\mathchar`-F1\mathchar`-measure$ from 0.66 to 0.76 on insufficient data.
\section*{Acknowledgment}
This work was supported in part by JSPS KAKENHI Grant Numbers JP19K12266, JP22K18006.
\bibliographystyle{IEEJtran}
\bibliography{anda}

% Generated by IEEJtran.bst, version: 0.17 (2021/08/26)
\begin{thebibliography}{1}
\providecommand{\url}[1]{#1}
\csname url@samestyle\endcsname
\providecommand{\newblock}{\relax}
\providecommand{\bibinfo}[2]{#2}
\providecommand{\BIBentrySTDinterwordspacing}{\spaceskip=0pt\relax}
\providecommand{\BIBentryALTinterwordstretchfactor}{4}
\providecommand{\BIBentryALTinterwordspacing}{\spaceskip=\fontdimen2\font plus
\BIBentryALTinterwordstretchfactor\fontdimen3\font minus
  \fontdimen4\font\relax}
\providecommand{\BIBforeignlanguage}[2]{{%
\expandafter\ifx\csname l@#1\endcsname\relax
\typeout{** WARNING: IEEJtran.bst: No hyphenation pattern has been}%
\typeout{** loaded for the language `#1'. Using the pattern for}%
\typeout{** the default language instead.}%
\else
\language=\csname l@#1\endcsname
\fi
#2}}
\providecommand{\BIBdecl}{\relax}
\BIBdecl

\bibitem{Lopez}
V.~L\'{o}pez, at~el.: ``{A}n {I}nsight into {C}lassification with {I}mbalanced
  {D}ata: {E}mpirical {R}esults and {C}urrent {T}rends on {U}sing {D}ata
  {I}ntrinsic {C}haracteristics'', Information Sciences, Vol.250, pp.113--141
  (2013)

\bibitem{BERT}
J.~Devlin, at~el.: ``{BERT}: {P}re-training of {D}eep {B}idirectional
  {T}ransformers for {L}anguage {U}nderstanding'', NAACL-HLT 2019,
  pp.4171--4186.

\bibitem{WordNet}
H.~Isahara, F.~Bond, K.~Uchimoto, M.~Utiyama, K.~Kanzaki: ``{D}evelopment of
  the {J}apanese {W}ord{N}et'', LREC 2008, pp.2420--2423.

\bibitem{Haque}
T.~U. Haque, N.~N. Saber, F.~M. Shah: ``{S}entiment {A}nalysis on {L}arge
  {S}cale {A}mazon {P}roduct {R}eviews'', ICIRD 2018, pp.1--6.

\bibitem{Xu}
H.~Xu, B.~Liu, L.~Shu, P.~Yu: ``{BERT} {P}ost-training for {R}eview {R}eading
  {C}omprehension and {A}spect-based {S}entiment {A}nalysis'', NAACL-HLT 2019,
  pp.2324--2335.

\bibitem{kobayashi2}
K.~Nozomi, at~el.: ``{E}xtracting {A}spect-evaluation and {A}spect-of
  {R}elations in {O}pinion {M}ining'', EMNLP-CoNLL 2007, pp.1065--1074.

\bibitem{EDA}
J.~Wei, K.~Zou: ``{EDA}: {E}asy {D}ata {A}ugmentation {T}echniques for
  {B}oosting {P}erformance on {T}ext {C}lassification {T}asks'', EMNLP-IJCNLP
  2019, pp.6382--6388.

\bibitem{Ekobayashi}
K.~Nozomi, I.~Kentaro, M.~Yuji, T.~Kenji, F.~Toshikazu: ``{C}ollecting
  {E}valuative {E}xpressions for {O}pinion {E}xtraction'', Journal of natural
  language processing, Vol.12, No.3, pp.203--222 (2005)

\end{thebibliography}
\end{document}